# Face Generation from Textual Features using Conditionally Trained Inputs to Generative Adversarial Networks

Sandeep Shinde[1], Tejas Pradhan[2], Aniket Ghorpade[3], Mihir Tale[4]

*Department of Computer Engineering, Vishwakarma Institute of Technology, Pune*

```
[1]sandeep.shinde@vit.edu
[2]tejas.pradhan18@vit.edu
[3]aniket.ghorpade18@vit.edu
[4]mihir.tale18@vit.edu
```

*Abstract*— Generative Networks have proved to be extremely effective in image restoration and reconstruction in the past few years. Generating faces from textual descriptions is one such application where the power of generative algorithms can be used. The task of generating faces can be useful for a number of applications such as finding missing persons, identifying criminals, etc. This paper discusses a novel approach to generating human faces given a textual description regarding the facial features. We use the power of state of the art natural language processing models to convert face descriptions into learnable latent vectors which are then fed to a generative adversarial network which generates faces corresponding to those features. While this paper focuses on high level descriptions of faces only, the same approach can be tailored to generate any image based on fine grained textual features.

## I. Introduction

Generative Adversarial Networks (GANs) have gained a lot of popularity owing to their state of the art performance in generating realistic images in very little time. GANs have been used for various applications including style transfer, improving image resolution, image restoration and deep fakes as well. The adversarial training realm of GANs has made it very difficult for humans to differentiate synthetic data samples from real ones. One such application of GANs is face generation. Many variants of GANs have been used to generate realistic looking fake human faces. Of these, NVIDIA's StyleGAN outperforms most other models in terms of fine grained details of faces [1]. These faces are generated using random vectors.

Deep learning methods for natural language processing applications have also reached state of the art performance in the recent past. Creating meaningful numerical representations from sentences and documents is an important task in the NLP space [2]. Models like Word2Vec and BERT assist in creating contextual embedding vectors which can then be used for a variety of machine learning tasks.

This paper proposes an approach which combines the power of natural language processing and generative adversarial networks to generate faces based on descriptions given to a neural network. Text is given as an input to the neural network which in the first phase converts it into appropriate word embeddings / numerical representations. This intermediate representation is an input to the GAN which will generate faces. Training such a neural network is possible using supervised learning techniques[3]. Our work introduces an architecture where a pre-trained word vectorization model can be made learnable by adding feed-forward layers to it. The loss between the expected face generated and the result of the GAN is directly back propagated to the vectorization model which learns appropriate numeric representations to generate faces in order to achieve the desired face output.

## II. Literature review

Generative adversarial networks for deep generative modelling were introduced by Goodfellow et. al. in the year 2014 which showed that the adversarial training mechanism of GANs proves to be efficient and accurate [4]. In 2016, Oord et. al. in their paper Conditional Image Generation with PixelCNN decoders implemented a conditional generative model using GANs which could generate images according to certain inputs [3]. They showed that Convolutional Neural Networks (CNNs) can be used to transform inputs into contextual word embeddings which are then fed as inputs to the GAN. Karras et. al. proposed a generative adversarial network in the year 2018, in their paper StyleGAN: A style based generator architecture for generative adversarial networks. StyleGAN has a number of different generator architectures which includes human faces as well. A pre-trained styleGAN can be used to generate realistic faces from random input latent vectors of a given size [1].

Chen X. et. al. explore a different approach to training face generation models given textual features in their paper FTGAN: A full trained generative adversarial network for text to face generation. The paper proposes an encoder decoder framework for training such models. The paper mentions that dividing the problem into text encoding and image decoding significantly improves the performance [5]. A similar approach was used by Nasir O. et. al. in their paper Text2FaceGAN Face generation from fine-grained textual features where they trained a deep convolutional generative adversarial network (DCGAN) with latent vectors as inputs [6]. The concept of DCGANs was introduced by Radford et. al. in their paper Unsupervised Representation Learning with Deep Convolutional Generative Adversarial Networks where they show that convolution layers can be used for generative modelling of synthetic data[7]. The most recent contribution on this topic is by Oza M. et. al. in their paper $ST^2GAN$: semantic text to face generative adversarial networks. They propose that BERT embeddings with attention layers work best to generate inputs for the GAN from textual features [8]. A very straightforward yet interesting strategy was used by Mirza et. al. in their paper titled conditional generative adversarial nets where they feed the data along with the expected label to the generator as well as discriminator of the GAN. This combination shows significantly good results in generating digits similar to the MNIST dataset given the digit to be generated [10].

Significant amount of work has been done in the area of generating numerical representations from text. The paper Efficient Estimation of Word Representations in Vector Space by Mikolov et. al. uses a method called Word2Vec which computes numerical representations of words in vector spaces by preserving similarity between words. Similar words have similar embeddings which are essential in mapping text features to numbers [2]. Transformer models with their attention mechanism have outperformed traditional vectorisation methods in the recent past as shown in the paper Attention is all You Need by Vaswani et. al [11]. Devlin et. al. in their paper Pre-training of Deep Bidirectional Transformers for Language Understanding propose a transformer based framework for contextual word embeddings [12]. An addition to this has been proposed by Reimers et. al. in their paper Sentence-BERT where they calculate sentence level embeddings using siamese networks to preserve semantic similarity [13].

Although the existing systems are able to synthetically generate faces using text inputs, they focus completely on training the GAN. However, research shows that training GANs is computationally complex and extremely difficult [16]. Secondly, it is difficult to quantify accuracy metrics by using image to image comparison. Our work introduces a new approach for this task by modelling it as a neural network regression problem and not having to train the GAN at all thereby making the process simpler, faster and more interpretable.

### III. DATA GENERATION

Since there is no readily available dataset to solve this problem, synthetic data had to be created. The input features for the data include textual descriptions of faces and the output for the same would be corresponding faces. A simple approach using this data was used in the first stage of research. Input text was fed to a neural network which would then generate face images with its shape depending on the shape of the neural network's output layer. The loss function used here was the mean of the pixel-wise difference between the original image (output label) and the generated image. However, this loss failed to provide a good convergence. One reason for this is that this approach focuses too much on generating the exact replica of the output image instead of focusing on tuning the face features.

A better approach to generating synthetic data which would solve this problem of loss was then used using StyleGAN. A block diagram of the complete data generation cycle is shown in Fig. 1. Around 20,000 random face images were generated by using StyleGAN by feeding random input vectors to it. StyleGAN is extremely efficient in creating realistic faces. The random input vectors fed to StyeGAN were also stored. The generated images were passed through an image captioning model trained on the ffhq (celebA faces) dataset. Fig. 2 shows the layerwise architecture of the image captioning model.It mainly consists of LSTM units which are used to generate text with image encodings given as an input. The Xception model trained on the imagenet dataset is used to generate the image encodings [14]. The result is 3 important pieces of information :

1. Input vectors for styleGAN.
2. Faces Corresponding to Input Vectors.
3. Descriptions corresponding to each face.

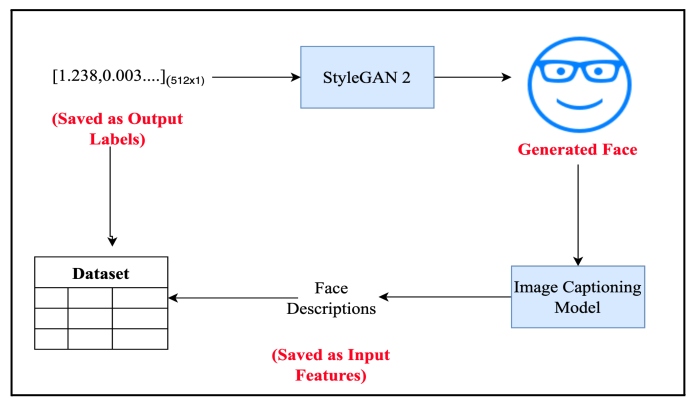

Fig. 1 Block Diagram of Automatic Dataset Generation

Here, the input to the neural network are the descriptions (text). However, the output of the network would be tuned vectors. These vectors would be then fed to styleGAN as inputs to generate faces. It is important to note that for a unique input vector, styleGAN generates a unique face. Modelling the problem in this manner has two advantages. Firstly, we no longer need to train the GAN. We can use a pre-trained generative model and train the inputs to the model instead which is far faster and easier than training the GAN. The reason is - this can be modelled as a regression problem by introducing the mean squared error between the output generated by the neural network and the input vector which generated the face corresponding to the text, as a loss function, thus making the training easier. Secondly, if we consider that each element in the input vector to the GAN contains some inherent facial features embedded in it, the neural network is essentially mapping the textual descriptions to its corresponding numerical face feature representation. This is exactly similar to embedding models like Word2Vec where a distributional semantic representation of each word is used. Here, the word vector is essentially based on the meaning of the word. Considering these points, the dataset generated contained the above mentioned 3 types of features with the text being the input for the neural network and numerical vectors being the output.

map the vectors to a particular face feature. These feature vectors will be fed as an input to a generative adversarial network which will be the neural network at the second stage. Concatenating these two networks and training them one after the other would generate conditional embeddings which can further be fed to the generative model to generate faces. The loss from the training of the generator can also be used to make the embeddings more accurate. Complete details regarding the training of stage-1 and stage-2 neural networks have been discussed further in the paper. Fig. 3 shows the overview of the proposed network and how the textual descriptions would be transformed to generate faces.

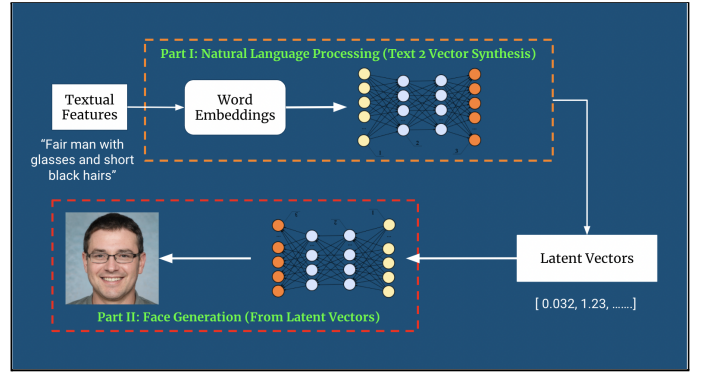

Fig. 3  Pipeline for Face Generation from Text

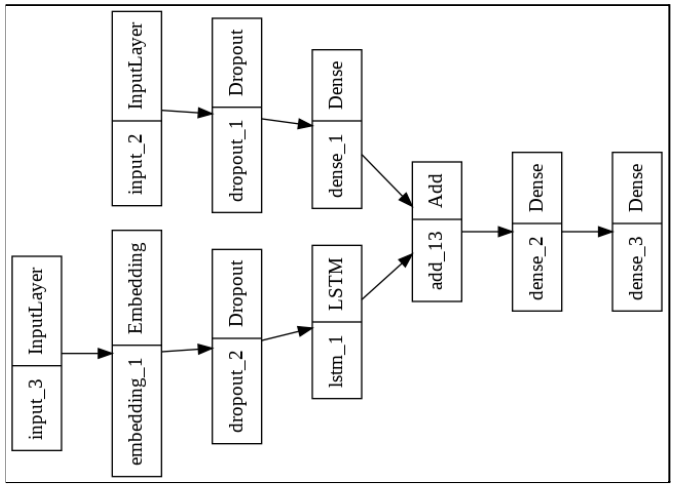

Fig. 2  Face Descriptor -Image Captioning Model Architecture

## IV. FACE GENERATION NEURAL NETWORKS

The proposed system consists of a two–staged neural network architecture. The first neural network would be used for converting the text descriptions into numerical feature vectors. The reason behind using a learnable network here is that the vector representation which is generated from the textual features should be such that the generative model can

### A. Text to Vector Synthesis using Sentence Transformers

The first stage in generating faces from textual features is to create learnable numerical vector representations from the text as shown in Fig. 3. In order to preserve only those words which are relevant to the facial features e.g. black eyes, double chin etc. the description has to be cleaned and preprocessed. This is done by removing stop-words and lemmatizing the terms so as to reduce the learning vocabulary for the generative neural network in the further stages. This text corpus can now be vectorised.

Several methods exist to generate vectors from words. Transformers have been an effective mechanism for generating vector representations while preserving features and word dependencies in text [11]. A pre-trained BERT transformer can be used to calculate word embeddings for the given text features. BERT stands for Bi-directional Encoder Representations from Transformers. The current implementation of our paper uses the uncased-base version of BERT which has 12 encoder layers, 12 attention heads and 110 million parameters. BERT has been pre-trained on a large text corpus. Its bi-directionality in transforming text into vectors is helpful in generating contextual embeddings. Fig.4.1 shows how an example sentence gets encoded through BERT.

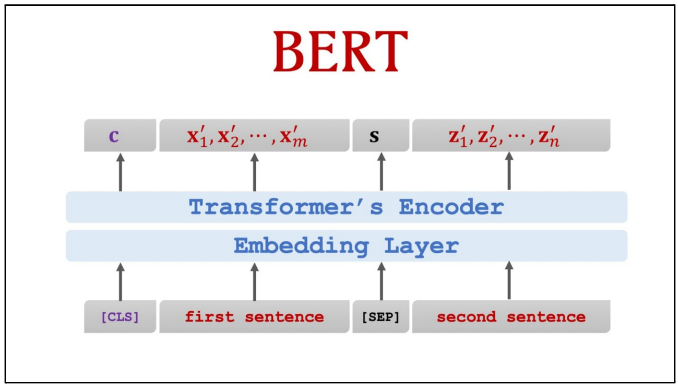

Fig. 4.1 BERT Transformer Architecture [12]

The use of transfer learning thus, reduces the training overhead for generating text features which preserve the word relations and context [12]. This, however, is not sufficient to generate semantically meaningful sentences which would be essential for reproducibility of the faces generated. Hence, we use the BERT sentence transformer for this task which has proved to be effective in calculating meaningful sentence embeddings. S-BERT uses a siamese neural network architecture similar to BERT for calculating word embeddings [13]. Siamese networks typically use 2 deep neural networks which share their weights. To preserve similarity, a distance metric is used for the output which is then passed to an objective function (transformer in case of BERT). Fig. 4.2 shows the general architecture of a Siamese neural network.

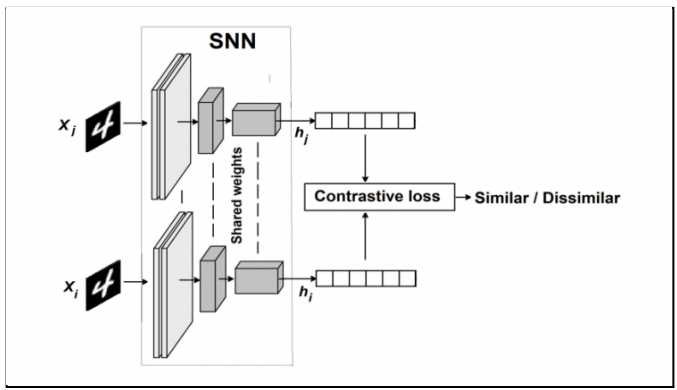

Fig. 4.2 Siamese Neural Network [15]

### B. Generating Learnable Vectors from Word Embeddings

The second stage in the architecture is to generate numeric vectors from the input word embeddings which can then be fed to the styleGAN to generate faces. For this purpose a feed-forward neural network consisting of 1-D convolution, 1-D max pooling and fully connected layers was used. Fig. 4.3 shows the layerwise architecture of the latent vector generation model. As discussed in section III, the use of such a feed forward representation would simplify the training process and abstract out a number of complications from the problem, thereby increasing the interpretability of the model. The output of this network was a 512x1 size vector which would be fed directly as an input to the styleGAN. The dimensions of the output layer are purposefully decided according to the input accepted by styleGAN.

Fig. 4.4 shows the final concatenated architecture of both the models. The models colored green are pre-trained and the model in blue, i.e, the face vector generator needs to be trained.

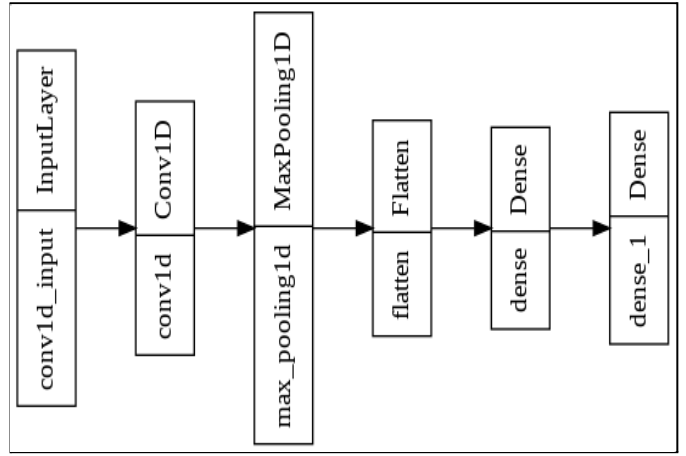

Fig. 4.3 Latent Face Vector Generator Model Architecture

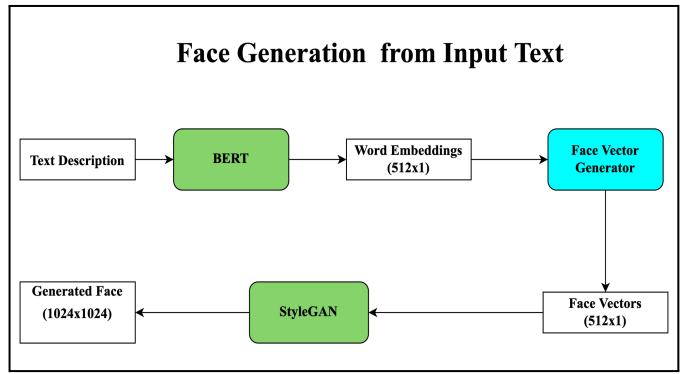

Fig. 4.4 Block Diagram of Stacked Face Generation Models

## V. NEURAL NETWORK TRAINING

Even though there are 3 distinct models being used in the entire pipeline, namely, BERT, feed-forward model, styleGAN; only one model needs to be trained. BERT and styleGAN are pre-trained and do not need any training. For the feed-forward model, a dataset of 20,000 images and corresponding descriptions was synthetically generated as discussed in section III.

The vector generation model was trained on 15,000 images for a total of 1200 epochs with 32 sentences per batch. 5000 images were used as the test set. The loss metric used was Mean Squared Error and the optimizer used was Adaptive Moment Estimation (Adam).

After 1200 epochs, the model showed a loss of 0.8819 MSE on the test set. Fig. 5.1 and 5.2 show the loss graphs during training. As shown in Fig. 5.1, the loss curves are smooth and the loss is decreasing at a steady rate. Although the loss shows a decreasing curve throughout the training, it can be clearly seen that there is no significant decrease in the loss during the final 400 epochs as in Fig. 5.2.

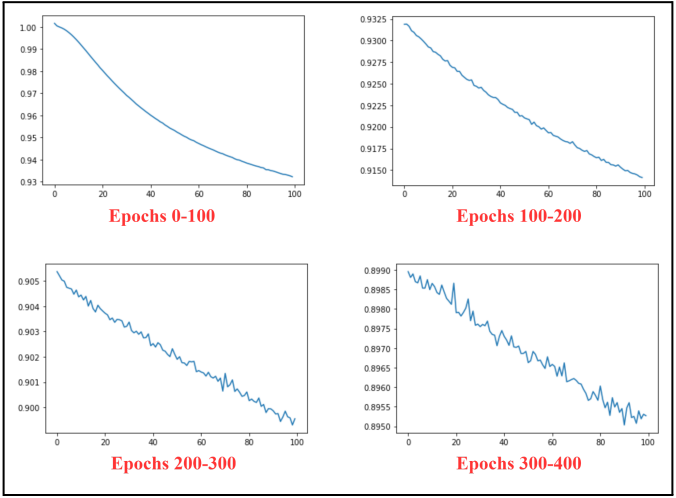

Fig. 5.1 Graph of training loss for the first 400 epochs

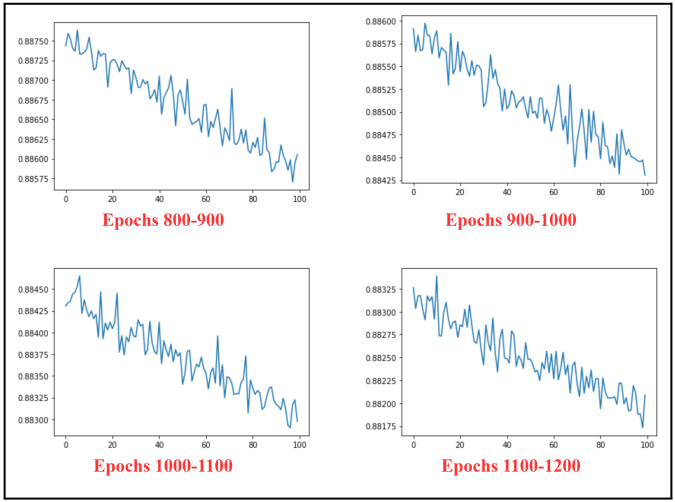

Fig. 5.2 Graph of training loss for the final 400 epochs

## VI. RESULTS

To observe how well the model has performed, merely using MSE as a metric is not enough. As the end goal of this paper is to generate realistic and accurate faces which correspond to descriptions given by users, it is important to understand how the faces are being generated. Many tests were carried out on the final model using the same pipeline discussed in the previous section to understand the quality of the output as well as increase the interpretability of our work. A detailed analysis and discussion of some of our results is given in this section.

**Input 1 :** "a young man with short dark hair and small dark eyes his lips are thin and his upper teeth are visible he is smiling a stubble beard is growing on his face"

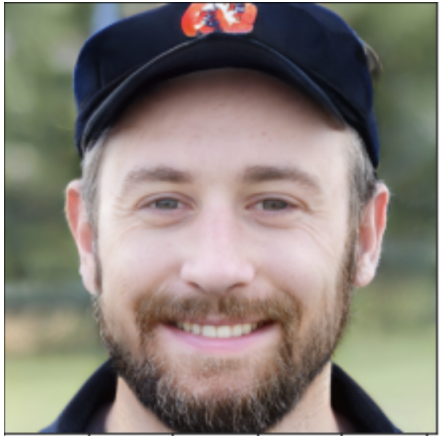

Fig. 6.1 Output Face Generated for Input 1

Fig 6.1 shows the output for the above input 1. From the image, it is clear that the model is able to distinguish between different genders. It is also observed that the model is able to understand finer details like "young" and "old". The image generated has very few wrinkles and is that of a young man as was needed. A smile with teeth visible is also seen in the generated face. From this, it can be inferred that the model is able to understand details regarding emotions like smiling, crying, laughing etc. A stubble beard is also generated accurately. However, in this image, the eye color was not captured correctly. The eye description was small and dark. Although the eyes appear small in the image, they are not dark in color. Another important point to note is that the model has itself introduced a cap on the man's head, even though it was not mentioned anywhere in the text. This means that the model is susceptible to some noise.

**Input 2 :** "a young boy with short wavy hair and a fringe he has a smile."

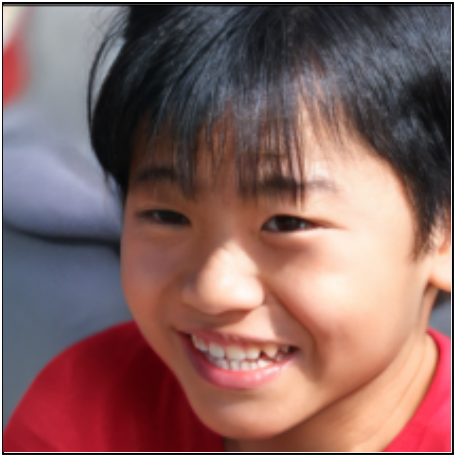

Fig. 6.2  Output Face Generated for Input 2

Fig. 6.2 shows the corresponding face generated for the above input text. It is clear that the hairstyle - fringes - is reflected in the image. Another important feature to note here is that the type of face generated for the first and second input changes significantly because the text here includes 'boy' instead of 'man'. It can be inferred from this that the model is able to understand age-related nuances and does not just focus on gender. In the first image, we can see a relatively young man but because of the presence of words like 'beard' the model outputs a face of an adult. For this input, however, the output face is that of a kid. This is mainly attributed to the text - 'boy' as discussed above. In addition to this, the expression on his face has also been captured from the text -'he has a smile'.

**Input 3:** "an old man with short grey hair and small dark eyes his lips are thin and he is smiling"

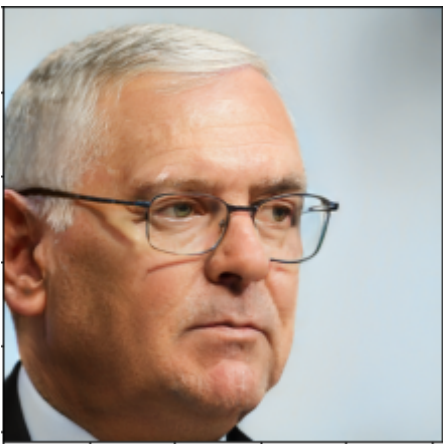

Fig. 6.3  Output Face Generated for Input 3

Fig. 6.3 shows the result of input text 3. This test case further strengthens the proposition of the model being able to create distinct faces and is able to understand the underlying meaning of the words such as 'young' , 'old' etc. Furthermore, in this specific case, we can see that even finer features like lip size have been modelled correctly. Hair length is also modelled correctly. The eye color, however, is not modelled correctly. The smile is also not that significant. A certain amount of tradeoff regarding which words to focus on can be observed here, wherein the model is focusing more on certain features and is oblivious to some even though they are modelled correctly in previous examples.

**Input 4:** "an old woman with short white hair and small dark eyes her lips are thin and she is smiling"

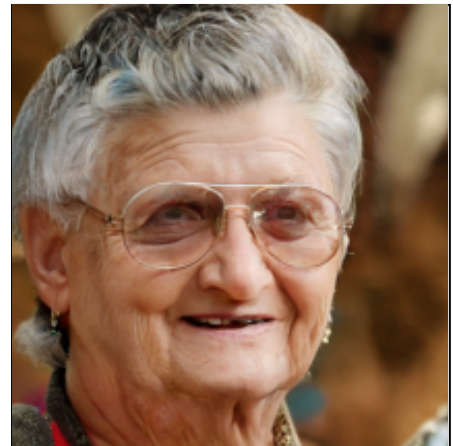

Fig. 6.4  Output Face Generated for Input 4

Fig. 6.4 shows the output face generated for input text 4. Two important features of the face that are correctly modelled in this image are the gender and age. We can clearly make out that this is the face of a woman. Additionally, there are features such as wrinkles which can be attributed to old age, given that the input text contained the word old. An interesting feature highlighted in the above image is the presence of earrings. This can be called noise, however, it is possible that the data which contained the word 'woman' had a lot of faces where women were wearing earrings. As a result the model learnt that the presence of earrings is also a feature attributed to the word woman. The hair color and hair style (short) have also been generated correctly. This is coherent with the previous results. From this, we can safely conclude that the model is able to understand the words related to length and color of hair to a great extent.

**Input 5:** "a sad woman with short white hair and a round face"

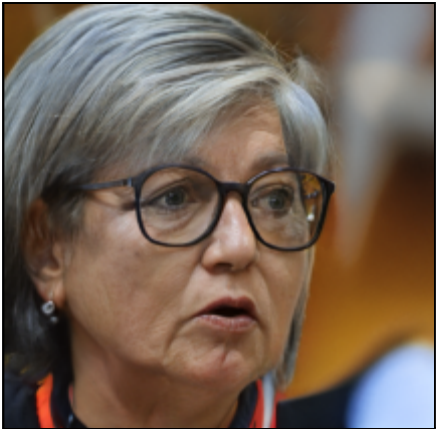

Fig. 6.5  Output Face Generated for Input 5

Fig. 6.5 shows the output face corresponding to the input text 5. The main feature tested here was emotions other than 'smile'. The addition of the word sad, generates a significantly different facial expression from that of smiling or happy. As in input 4, the short white hair have been modelled perfectly in this face. . Along with this, the round structure of the face is also clearly visible in this image. As a result, we can infer that not only is the model able to generate faces corresponding to certain facial expressions, but changing even a single word also significantly changes the emotion depicted. Hence, the model is able to generate faces corresponding to opposite emotions as well and is not biased to certain expressions.

**Input 6:** "a girl with blonde hair wearing dark shades and her lips are thin and she is sad"

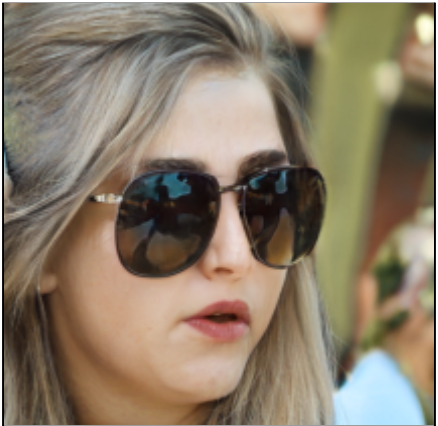

Fig. 6.6  Output Face Generated for Input

Fig. 6.6 shows the result for input text 6. In contrast to image 6.4 and 6.5, we can clearly see that the woman's face resembles someone who is very young. We can also see that the expression is that of a sad person. Another feature tested here is the inclusion of accessories. We could see certain accessories as noise but here, in this picture, we can clearly see that dark shades have been reflected in the face generated. The hair color is blonde which is as expected.

## VII. CONCLUSION

This paper discusses a novel approach to generating images, specifically faces using conditional inputs to generative adversarial networks. Instead of relying completely on the core accuracy of the GAN, which is known to be very difficult to train, our work focuses on training the inputs to the GAN and treating the GAN as a constant function. Even though we are able to generate high quality images from only a few input sentences, currently every unique input sentence given to the model has only one unique output. Having multiple output faces generated with slight variance with respect to features and surroundings can make the model even more robust. This can be implemented in future by adding some small random noise to the input (to GAN) and generating multiple different inputs that are very close to each other, these inputs when passed to the StyleGAN will give us similar faces as results but with some variation out of which the most suitable face can be selected. This process can be done iteratively to reach closer and closer to the most accurate face that the user wants to describe.